\pdfoutput=1
\documentclass[runningheads]{llncs}
\usepackage{bbding}
\usepackage[T1]{fontenc}
\usepackage{microtype}
\usepackage{graphicx}
\usepackage{subfigure}
\usepackage{tabularx}
\usepackage{booktabs} 
\usepackage{subcaption}
\usepackage{setspace}
\usepackage{hyperref}
\usepackage{amsmath}
\usepackage{amssymb}
\usepackage{mathtools}
\usepackage{algorithm}
\usepackage{algorithmic}
\usepackage{orcidlink}

\let\originalunskip\unskip
\renewcommand{\unskip}{}

\begin{document}
\title{Incomplete Multi-view Multi-label Classification via a Dual-level Contrastive Learning Framework}
%
%

\author{Bingyan Nie\inst{1}~\orcidlink{0009-0009-3556-6040}
\and Wulin Xie\inst{1}~\orcidlink{0009-0000-9535-4116} 
\and Jiang Long\inst{1}~\orcidlink{0009-0000-6048-0737} 
\and Xiaohuan Lu\inst{1(}\Envelope\inst{)}~\orcidlink{0009-0006-7634-7272} 
}
\authorrunning{B. Nie et al.}
%
\institute{\inst{1}School of Big Data and Information Engineering, Guizhou University, Guiyang 550000, China \\
\email{xhlu3@gzu.edu.cn}
 }

\maketitle             
\begin{abstract}

Recently, multi-view and multi-label classification have become
significant domains for comprehensive data analysis and exploration. However, incompleteness both in views and labels is still a real-world scenario for multi-view multi-label classification. In this paper, we seek to focus on double missing multi-view multi-label classification tasks and propose our dual-level contrastive learning framework to solve this issue. Different from the existing works, which couple consistent information and view-specific information in the same feature space, we decouple the two heterogeneous properties into different spaces and employ contrastive learning theory to fully disentangle the two properties. Specifically, our method first introduces a two-channel decoupling module that contains a shared representation and a view-proprietary representation to effectively extract consistency and complementarity information across all views. Second, to efficiently filter out high-quality consistent information from multi-view representations, two consistency objectives based on contrastive learning are conducted on the high-level features and the semantic labels, respectively. Extensive experiments on
several widely used benchmark datasets demonstrate that the proposed method has more stable and superior classification performance.

\keywords{Incomplete multi-view learning  \and Incomplete multi-label classification \and Contrastive learning.}
\end{abstract}
\let\unskip\originalunskip

\section{Introduction}
In recent years, multi-view data, including images, audio, text descriptions, and videos~\cite{Alpher01,Diao_2025_WACV}, not only expands swiftly but also has attracted lots of attention and is crucial for various applications such as cross-view retrieval~\cite{Alpher04}, sentiment analysis~\cite{Alpher05} and image annotation~\cite{Alpher06,diao2025temporal,diao-etal-2024-learning}. For example, for images, a variety of visual characteristics such as HOE, SIFT, RGB, GIST and LBP~\cite{Alpher07} can be obtained by conventional filtering techniques which together constitute a form of multi-view data. 

Multi-view classification (MVC) aims to assign samples to predefined categories based on their features~\cite{Alpher08,Alpher09}, generally focusing on using information from different views that contain the unique features of each view and the similar semantic structure among views. As a consequence, multi-view classification arises out of necessity, and a large number of approaches based on subspace learning ~\cite{Alpher11,Alpher12}, collective matrix factorization~\cite{Alpher15,Alpher16}, and label embedding ~\cite{Alpher18,Alpher19} have been proposed. In practice, the majority of samples may be assigned more than one label~\cite{Alpher21,Alpher22}, due to the diversity of assignment and the data being analyzed. Specifically, in text classification, a news report may be simultaneously labeled with multiple tags, such as “sports,” “entertainment,” “technology,” and “politics”. Thus, in the computer vision community, multi-label classification has become particularly important and attracted increasing research efforts. Insight of this, the scenario designated as multi-view multi-label classification (MVMLC) has attracted more attention than traditional tasks that focus on either multi-view or multi-label classification.

For MVMLC, researchers have carried out numerous methods in the past few years~\cite{Alpher24,Alpher26}. For instance, Liu et al integrate multiple feature views into a low-dimensional subspace, using a novel low-rank multi-view learning algorithm and matrix completion techniques~\cite{Alpher24}. Additionally, neural networks are also used to solve this issue. For example, a novel neural network multi-view multi-label learning framework (CDMM) is proposed, which is intended to solve the problem of consistency and diversity among views through a simple and effective method~\cite{Alpher26}. However, it also introduces a challenging yet important problem for MVMLC, due to the mistakes in manual annotation, which cause the partial absence of data~\cite{Alpher28}. To solve the problem, a deep instance-level contrastive network(DICNet) utilizes deep neural networks to extract high-level semantic representations and deploys an instance-level contrastive learning strategy to enhance the extraction of consistent information across different views~\cite{Alpher31}. Then, a novel incomplete multi-view partial multi-label learning (IMVPML) framework has been proposed to obtain ground-truth labels with a low-rank and sparse decomposition scheme and a graph Laplacian regularization~\cite{Alpher32}. To display own characteristics of each view and consistent representations of the same sample in multiple views, Liu et al proposed a masked two-channel decoupling framework (MTD) to decouple each view’s latent feature into two types of features ~\cite{Alpher33}. However, they disregard considering multi-label correlation and the conflict in learning target between the consistency objective and reconstruction objective, i.e., the consistency objective and the reconstruction objective are pushed on the same features which denote the latent features of raw data.

In this paper, we propose a dual-level contrastive learning framework (DCL for short) to address the aforementioned issues. Our purpose is to maximize the extraction of shared information across all views while preserving the unique information specific to each view. The aforementioned issues are challenging as many works attempt to identify the samples’ category by integrating the features from all views, in which process meaningless view-private information may overshadow the common semantics, and thus influence the quality of classification~\cite{Alpher34}. To address this, decoupling the latent feature of each view into two separate types to fully extract low-level features is needed~\cite{Alpher33}. In addition, we design two-level contrastive learning to avoid the consistency objective and reconstruction objective on the same features. Concretely, one is a multi-view contrastive learning method considering the relationship between different views and different samples, another is a multi-label contrastive learning method that aims to minimize the distance of correlational labels across all views. As a result, the conflict between the reconstruction objective and the two consistency objectives is alleviated. In summary, our main contributions are outlined as follows:

\begin{itemize}
\item We propose a novel dual-level contrastive learning (DCL) for the IMVMLC task. The proposed novel could restrain different learning objectives at different levels, reducing the conflict between the consistency objective and the reconstruction objective. Thus, our approach can discover the shared semantics across all views while preserving meaningful view-specific information.
\item Our incomplete instance-level contrastive learning focuses on extracting high-level semantic features, guiding the shared feature encoder to extract cross-view semantic features with better consensus. Simultaneously, in the supervised setting, labels are used to construct a wider variety of positive pairs from different views of the same class. As a result, representations learned in label-level contrastive learning are satisfactory for the downstream task based on supervisory labels.  
\item Sufficient experiments on five datasets demonstrate the effectiveness of a dual-level contrastive learning framework in solving the double-missing case.
\end{itemize}

\begin{figure}
\includegraphics[width=\textwidth]{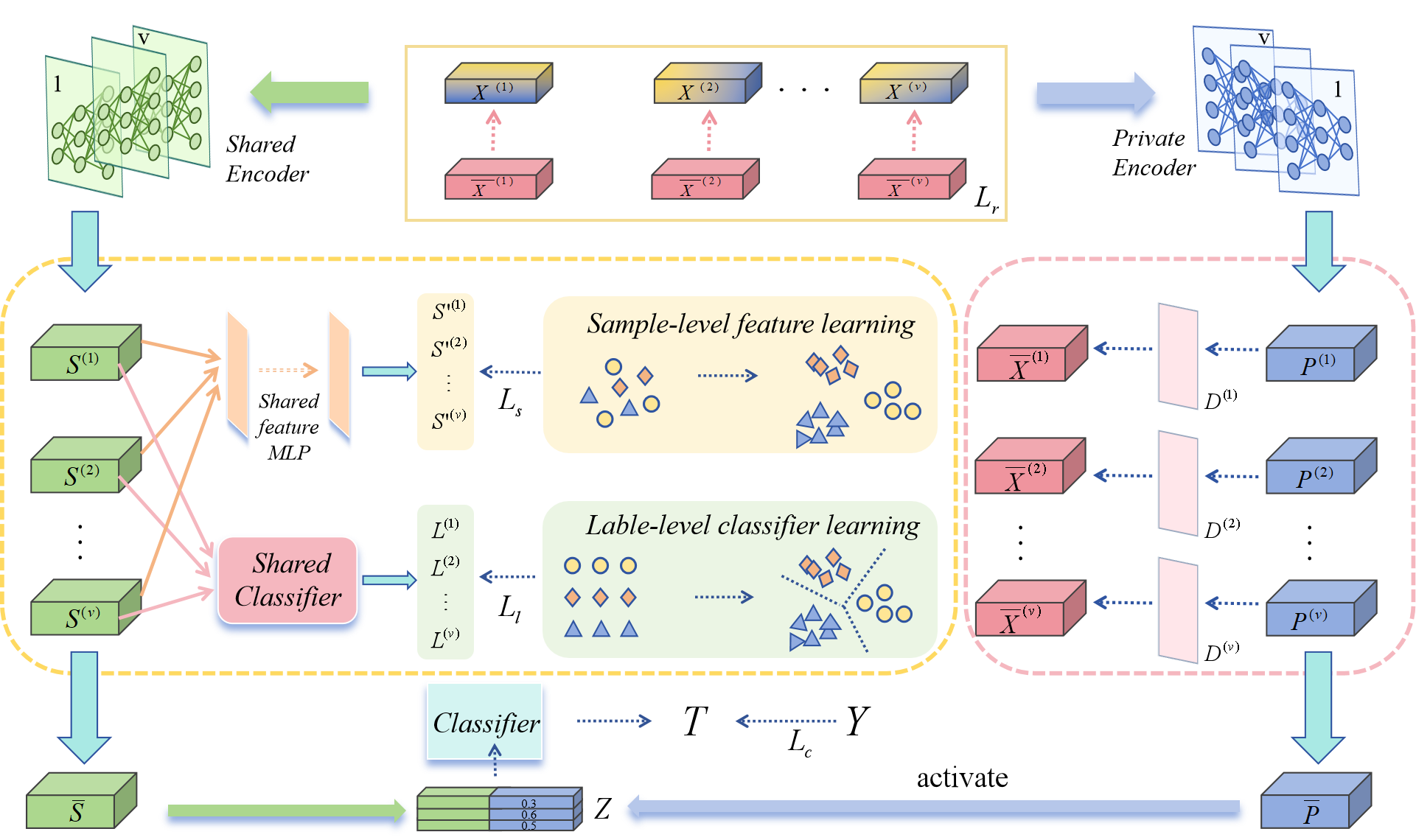}
\caption{The main framework of our DCL. The private and consistent features are extracted by S-P encoders, respectively. Adopting an interaction approach, $Z$ represents the final fused fusion.}
\label{fig1}
\end{figure}
\section{Related work}

\subsection{Incomplete Multi-view Multi-label Classification}
In recent years, the difficulties encountered in multi-view data collection have led to the possibility that some views may not contain complete information in real tasks. This may affect the performance of traditional partial multi-view learning algorithms. Consequently, the incomplete multi-view multi-label classification (IMVMLC) task has attracted a growing interest from numerous researchers and the corresponding methods have demonstrated remarkable efficacy. For instance, Li and Chen proposed a new model, NAIM3L, to explicitly model the global high-rank and local low-rank structures within multiple labels \cite{Alpher35}. To address the double missing problem, NAIM3L applied a prior missing indicator matrix including labels and views missing information. In addition, the contrastive network is employed, as the DICNet proposed by Liu et al., which aims to aggregate instances of the same sample across different views and segregate instances belonging to different samples \cite{Alpher31}. Additionally, Zhu et al. proposed a method named WCC-MVML-ID, which integrates within-view, cross-view, and consensus-view representations to effectively process incomplete multi-view multi-label datasets\cite{Alpher41}. Liu et al. proposed LMVCAT leverages the self-attention mechanism to extract high-level features and exploit inter-class correlations to enhance classification performance\cite{Alpher42}. Another framework in IMVMLC has also shown its remarkable performance, namely, MTD \cite{Alpher33}. Although MTD has designed a cross-channel contrastive loss and a graph constraint approach to maintain the structural information in learned embedding features, it still neglects the correlation of multi-label and missing data recovery. 
\subsection{Contrastive Learning}
Over the years, contrastive learning has demonstrated considerable potential in the field of supervised and unsupervised representation learning \cite{Alpher10,Alpher31,Alpher34,diao2025soundmind,diao2025learning}. The fundamental concept is to identify a concealed area where the consensus between different views of the same sample is maximized by contrasting the agreement across different samples. For example, in DICNet, Liu et al. devised an instance-level contrastive loss to direct the autoencoder to learn cross-view high-level features following the consensus hypothesis\cite{Alpher31}. Supervised contrastive (SC) learning \cite{Alpher36}, an extension of contrastive learning, integrates label information to generate positive and negative pairs. In contrast to conventional alternatives which combine an anchor and a “positive” sample in embedding space, and separate the anchor from numerous “negative” samples, the use of many positives and negatives for each anchor enables superior performance without the need for difficult negative mining \cite{Alpher36}, which can be challenging to optimize. 

\section{Method}
Inspired by the analysis of contrastive learning, in this section, we propose a new dual-level contrastive learning framework, whose main structure is shown in Figure 1. Our model possesses the following three aspects, namely, low-level decoupling features, instance-level contrastive learning and label-level contrastive learning. First of all, for the convenience of description, we brieﬂy outline the formal problem definition and frequently used notations.

\subsection{Formulation}
For an incomplete multi-view multi-label data set, there are $v$ data views, and each view of the data is represented by $\left \{ X^{(m)}\in \mathbb{R} ^{N\times d_{m} }   \right \} _{m=1}^{v} $, where $d_{m}$ is the dimensional of $m$-th view, and $N$ is the number of samples. Additionally, $Y\in \left \{ 0,1 \right \} ^{N\times C} $ is the label matrix, where $C$ is the number of categories. If $i$-th sample has $j$-th label, $Y_{ij} =1$ otherwise, $Y_{ij} =0$. Considering the missing views and missing labels, we introduce two missing matrices called the missing-view indicator and the missing-label indicator, respectively. And the former matrix is denoted as $V\in \left \{ 0,1 \right \} ^{N\times v} $, in which $V_{ij} =1$ means $j$-th view of $i$-th sample is available, otherwise $V_{ij} =0$. The latter matrix $W\in \left \{ 0,1 \right \} ^{N\times C} $ is similar to the missing-view indicator, where $W_{ij} =1$ indicates $j$-th category of sample $i$ is known. For simplicity, we fill the missing views and unknown labels with ‘0’ in the data-preparation stage. The task of DCL is to train a model which can appropriately predict multiple categories for each input sample with incomplete views and labels.

\subsection{low-level decoupling features}
As is known to all, the input raw data is low-quality, e.g., containing noisy and missing values, which is unsuitable to directly learn semantic information \cite{Alpher30,Alpher31,Alpher32,Alpher33}. Generally, autoencoder \cite{Alpher20,Alpher34} is a widely used unsupervised model that can project the raw features into a customizable feature space. Inspired by \cite{Alpher33}, it applies the masked autoencoders (MAE) that randomly mask patches of the input image. By employing this method, we produce a matrix $M^{(m)} \in \left \{ 0,1 \right \} ^{N\times d_{m} } $ for each view whose elements are initialized to 1. Then we select $N$ integers in random as the masking start of each sample in $m$-th view, after which 1 is filled with 0. Finally, we can achieve the same feature dimension of each masked view, denoting as:
\begin{equation}
\left \{ X'^{(m)} \right \}_{m=1}^{v} =\left \{  X^{(m)} \otimes M^{(m)}  \right \} _{m=1}^{v} 
\end{equation}
in Eq(1), $\otimes$ means the element-wise multiplication. Furthermore, data from different views usually contain consistent information and single-view personal information. To this end, we introduce S-P decoupling encoders for each view to extract the low-level features, where $\left \{ E_{(m)}^{S}:X'^{(m)} \to S^{(m)}   \right \}_{m=1}^{v}$ and $\left \{ E_{(m)}^{P}:X'^{(m)} \to P^{(m)}   \right \}_{m=1}^{v}$ represent the shared encoder and private encoder, respectively. Simultaneously, $S^{(m)} \in \mathbb{R} ^{N\times d} $ and $P^{(m)} \in \mathbb{R} ^{N\times d}$ are the extracted consistent feature matrix and view-complementary feature matrix for each view of $N$ samples. On the one hand, shared encoder $\left \{ E_{(m)}^{S}  \right \}_{m=1}^{v}  $ can learn a more discriminative consensus representation and private encoder $\left \{ E_{(m)}^{P}  \right \}_{m=1}^{v}$ can remain the structure of initial data. On the other hand, designing S-P encoders alleviates the conflict between the consistency objective and the reconstruction objective, conducting them into two latent feature spaces, where the consistency objective is achieved by the following dual-level contrastive learning. So the reconstruction objective of each view is formulated as:
\begin{equation}
L_{r} =\frac{1}{v} \sum_{m=1}^{v} \ell _{r}^{(m)} =\frac{1}{v} \sum_{m=1}^{v} \sum_{i=1}^{N} \frac{1}{d} \left \| \bar{X}_{i}^{(m)} - X_{i}^{'(m)}    \right \|_{2}^{2}V_{im}
\end{equation}
where $\bar{X} ^{(m)}$ is the output of decoder $\left \{ D_{m}:P^{(m)}\in \mathbb{R}^{N\times d}  \to \bar{X}^{(m)} \in  \mathbb{R}^{N\times d}  \right \}_{m=1}^{v}$ and only when $V_{im}$  is available, $m$-th view of $i$-th sample will be calculated. Furthermore, the reconstruction loss between input $X^{'(m)}$ and output $\bar{X} ^{(m)} $ is defined as $\ell _{r}^{(m)} $, and $L_{r} $ denotes the average reconstruction loss among all views. 

\subsection{instance-level contrastive learning}
Since the latent features $P^{(m)} \in \mathbb{R} ^{N\times d} $ and $S^{(m)} \in \mathbb{R} ^{N\times d} $ obtained by S-P decoupling model mix the semantics with the view-consistency and the view-proprietary information, we treat them as low-level features. Considering that contrastive learning plays an important role in helping encoders extract more common contents \cite{Alpher10,Alpher31,Alpher34}, we design a shared feature MLP to learn another level of features, i.e., instance-level features. Then the $m$-th view of $\left \{ S^{'(m)}  \right \}_{m=1}^{v} $ is represented as $S^{'(m)} \in \mathbb{R} ^{N\times d}$ in the high-level feature space, where $S'$ indicates the sample-level assignment .
Specifically, for each sample $S_{i}^{'(m)} $ learned from the high-level space, we remark it as an anchor instance while it has vN-1 instance pairs, i.e., $\left \{ S_{i}^{'(m)},S_{j}^{'(n)}  \right \} $. In addition, there are (v-1) positive pairs which belong to one sample but not in a view, such as $\left \{ S_{i}^{'(m)},S_{i}^{'(n)}  \right \}_{m\ne n}  $ and the rest v(N-1) feature pairs are negative feature pairs of each high-level feature. It is our motivation that devotes to minimizing the distance of available negative pairs and maximizing the similarity of available positive instance-pairs. So we adopt cosine distance to measure the similarity of instance-pairs:
\begin{equation}
d\left (  S_{i}^{'(m)},S_{j}^{'(n)}\right ) = \left ( \frac{\left \langle S_{i}^{'(m)},S_{j}^{'(n)}  \right \rangle }{\left \| S_{i}^{'(m)} \right \|\left \| S_{j}^{'(n)} \right \|  } +1 \right )/2
\end{equation}
where $\left \langle \cdot ,\cdot  \right \rangle $ is dot product operator, and our optimization purpose is to reduce the $d\left (  S_{i}^{'(m)},S_{j}^{'(n)}\right )$. Noted that the view of $S^{'} $ is partial missing, therefore the missing-view matrix is introduced to measure the instance-level contrastive loss:
\begin{equation}
\ell _{sc}^{mn} = -\frac{1}{N} \sum_{i=1}^{N} V_{im} V_{in} \log_{}{\frac{e^{d\left (  S_{i}^{'(m)},S_{i}^{'(n)}\right )/\tau _{s} } }{\left (  {\textstyle \sum_{j=1}^{N}} {\textstyle \sum_{k=m,n}^{}} e^{d\left (  S_{i}^{'(m)},S_{j}^{'(k)}  \right )/\tau _{s}  } V_{jk}  \right )-e^{1/\tau _{s}  }  } }
\end{equation}
where $ \tau $ represents the temperature parameter that regulates the concentration extent of the distribution and $\ell _{sc}^{mn}$ indicates the contrastive loss between $S^{'(m)} $ and $S^{'(n)} $. We introduce $V_{im} V_{in}$ to mask missing views in the process of calculation loss $\ell _{sc}^{mn}$. The overall high-level feature contrastive loss across all view pairs is shown as:
\begin{equation}
L_{s} = \frac{1}{2} \sum_{m=1}^{v} \sum_{n\ne m}^{} \ell _{sc}^{(mn)}  
\end{equation}
As can be seen from Eq.(4), the cross-entropy loss of $i$-th sample in relation to $ m$-th view and $n$-th view will only be calculated when the positive instance pair $ (S_{i}^{'(m)},S_{i}^{'(n)})$ are both available.

\subsection{label-level contrastive learning}
It is well known that exploiting label correlations has a crucial impact on multi-label classification \cite{Alpher37}. Additionally, learning it can effectively reduce the number of labels needed to be predicted and optimize the classification performance. Following other multi-label classification works \cite{Alpher37,Alpher38}, our model utilizes a label MLP on shared information $\left \{ S^{(m)}  \right \}_{m=1}^{v} $, where the prediction of each view is regarded as $n$ independent binary classification problems. Similar to learning the sample-level features, the label-level objective is represented as $\left \{ L_{i} ^{(m)}  \right \}_{m=1}^{v} $, where $L^{(m)} \in \left [ 0,1 \right ]^{N\times C}  $ implies the category results of $m$-th perception. According to Eq(3), we leverage cosine distance to measure the similarity across all labels:
\begin{equation}
d\left ( L_{i}^{(m)}, L_{j}^{(n)} \right ) = \left ( \frac{\left \langle L_{i}^{(m)},L_{j}^{(n)} \right \rangle }{\left \| L_{i}^{(m)} \right \|\left \| L_{j}^{(n)} \right \|  } +1 \right )/2 
\end{equation}
Here, we regard $L_{i}^{(m)} \in \left [ 0,1 \right ] ^{1\times C} $  as semantic features in label space of sample $i$ in view $m$ where $C$ denotes the dimension of semantic features. As a result, we mark all features in  $\left \{ L^{(m)}  \right \} _{m=1}^{v} $ with three types: (1) \textbf{anchor example} $L_{i}^{(m)}$ (2) \textbf{negative pairs} $\left \{ L_{i} ^{(m)},  L_{j} ^{(n)}\right \} _{i\ne j}$ (3) \textbf{positive pairs} $\left \{ L_{i} ^{(m)},  L_{i} ^{(n)}\right \} _{n\ne m}$. Similar to instance-level contrastive learning, we further define the label-level contrastive loss between $L^{(m)}$ and $L^{(n)}$ as:
\begin{equation}
\ell _{lc}^{mn} = -\frac{1}{N} \sum_{i=1}^{N} W_{im} W_{in} \log_{}{\frac{e^{d\left (  L_{i}^{'(m)},L_{i}^{'(n)}\right )/\tau _{l} } }{\left (  {\textstyle \sum_{j=1}^{N}} {\textstyle \sum_{k=m,n}^{}} e^{d\left ( L_{i}^{'(m)},L_{j}^{'(k)}  \right )/\tau _{l}  } V_{jk}  \right )-e^{1/\tau _{l}  }  } }
\end{equation}
where $\tau$ represents the temperature parameter and $W_{im} W_{in}$ is introduced to ignore the missing labels during calculation. As a result, the total label-level learning objective is defined as:
\begin{equation}
L_{l} = \frac{1}{2}\sum_{m=1}^{v}\sum_{n\ne m}^{}\ell _{lc}^{(mn)}     
\end{equation}
From aforementioned two-level contrastive learning method, $\left \{ S^{'(m)}  \right \} _{m=1}^{v}$ and $\left \{ L^{(m)}  \right \} _{m=1}^{v} $ would achieve the cross-view consistency while preserving the view-specific complementary information in $\left \{ P^{(m)}  \right \} _{m=1}^{v}$. Similar to previous fusion strategies, it is natural to get cross-view fusion to obtain the unique shared features $\bar{S} $ and $\bar{P} $ of all examples without the negative effects of missing views:
\begin{equation}
\bar{S} _{i} =\sum_{m=1}^{v} \frac{S_{i}^{(m)}V_{im}  }{ {\textstyle \sum_{m}^{ }V_{im} }},  \bar{P} _{i} =\sum_{m=1}^{v} \frac{P_{i}^{(m)}V_{im}  }{ {\textstyle \sum_{m}^{ }V_{im} }}
\end{equation}
where $ \sum_{m}^{ }V_{im} $ represents the number available among $v$ views in $i$-th sample. Then we adopt an interaction approach to fuse shared features and view-specific features:
\begin{equation}
Z_{ij} = \theta \left ( \bar{P}_{ij} \right )  \cdot \bar{S}_{ij}  
\end{equation}
where $Z\in \mathbb{R} ^{N\times d} $ represents the final fused feature of $i$-th sample and $m$-th view, combining private information $\bar{P}_{ij}$ and consistent information $\bar{S}_{ij}  $, and $\theta$ denotes the sigmoid activation. To further enhance the classification performance, label matrix $Y\in \left \{ 0,1 \right \}^{N\times c} $ is used to guide the prediction result of final fused feature $Z\in \mathbb{R} ^{N\times d}$ :
\begin{equation}
T = Sigmoid\left ( Z\omega +  \lambda  \right ) 
\end{equation}
where $T \in \mathbb{R}^{N\times c}$ denotes the predicted score matrix that $Z$ maps to in the label space via a classifier. Besides, the $\omega \in \mathbb{R} ^{d\times c}  $ and $\lambda  \in \mathbb{R} ^{N\times c} $ belong to the learnable parameters of classifier. Further, we let label matrix $Y \in \left \{ 0,1 \right \}^{N\times c}$ be the target and prediction result be the learning object:
\begin{equation}
L_{c} = -\frac{1}{N} \frac{1}{c} \sum_{i=1}^{N}\sum_{j=1}^{c} \left ( Y_{ij} \log_{}{\left ( T_{ij}  \right ) +\left ( 1-Y_{ij}  \right )\log_{}{\left ( 1-T_{ij}  \right ) }  }  \right )W_{ij}
\end{equation}
where $W_{ij}$ is introduced to filter out missing tags in the calculation of $L_{c}$. Taking these four losses together, the total loss of our DCL model can be expressed as:
\begin{equation}
L_{all} = L_{c + } \alpha L_{s} +  \beta L_{l} + \gamma L_{r}   
\end{equation}
where $\alpha ,\beta ,\gamma $ are penalty parameters with respect to $L_{s},L_{l},L_{r} $. In the period of training data, all parameters will be updated via backpropagation.

\begin{algorithm}[!t]
 \caption{Training process of \textbf{DCL}.}
  \begin{algorithmic}
  \STATE \textbf{Input}: Incomplete multi-view data $\left \{ X^{(m)} \right \}_{m=1}^{v}$ with its corresponding missing-view indicator matrix $V \in \left \{ 0,1 \right \}^{N \times v}$, and weak label $Y \in \left \{ 0,1 \right \}^{N \times C}$ with its related missing-label indicator matrix $W \in \left \{ 0,1 \right \}^{N \times C}$.
  \STATE \textbf{Initialization}: Fill unavailable elements with 0; Set $L_{all}=0$; Initialize model weights, hyper-parameters $\alpha, \beta, \gamma$, learning rate, and training epochs $E$.
  \STATE \textbf{Output}: Updated model parameters.
  \WHILE{$k < E$}
    \FOR{$m = 1$ \textbf{to} $v$}
      \STATE Compute $X'^{(m)} = X^{(m)} \otimes M^{(m)}$
    \ENDFOR
    \STATE Extract embedding features via shared encoder $\left \{ E_{(m)}^{S} \right \}_{m=1}^{v}$ and private encoder $\left \{ E_{(m)}^{P} \right \}_{m=1}^{v}$, respectively
    \STATE Compute $L_r$ by Eq.(2) using decoder $\left \{ D_{m} : P^{(m)} \to \bar{X}^{(m)} \right \}_{m=1}^{v}$
    \STATE Obtain instance-level features $\left \{ S^{'(m)} \right \}_{m=1}^{v}$ using a shared feature MLP
    \STATE Compute high-level contrastive loss $L_s$ by Eq.(5)
    \STATE Obtain label-level features $\left \{ L_{i}^{(m)} \right \}_{m=1}^{v}$ using a label MLP
    \STATE Compute label-level contrastive loss $L_l$ by Eq.(8)
    \STATE Compute cross-view fusion by Eq.(9), respectively
    \STATE Compute final fused features by Eq.(10)
    \STATE Obtain predictions $T$ by Eq.(11)
    \STATE Compute classification loss by Eq.(12) and total loss by Eq.(13)
  \ENDWHILE
  \end{algorithmic} 
\end{algorithm}

\section{Experiments}
This section presents the experimental setup and analysis employed to evaluate the proposed method in detail. 
\subsection{Experimental Setting}
\textbf{Datasets}: We evaluate our model on five popular multi-view multi-label datasets, following \cite{Alpher31,Alpher35,Alpher39}: (1) \textbf{Corel5k} (2) \textbf{Pascal07}  (3) \textbf{ESPGame} (4) \textbf{IAPRTC12} (5) \textbf{MIRFLICKR} The five datasets encompass a wide range of samples, from 4,999 to 25,000, and a corresponding range of categories, from 20 to 291. Additionally, six distinct types of features were selected as six views, namely GIST, HSV, DenseHue, DenseSift, RGB, and LAB.

\textbf{Compared Methods}: In our experiments, we compare the proposed \textbf{DCL} with six popular methods in the field, i.e., \textbf{CDMM} \cite{Alpher26}, \textbf{NAIM3L} \cite{Alpher35}, \textbf{iMVWL} \cite{Alpher39}, \textbf{DICNet} \cite{Alpher31}, \textbf{LMVCAT} \cite{Alpher28} and \textbf{MTD} \cite{Alpher33}, showing the advancement of our model on the five aforementioned datasets. All these methods have been previously discussed and these solutions address different tasks. Concretely, \textbf{CDMM} is designed to address the MVMLC task, however, without considering the missing situation. Furthermore, the IMVMLC task is our important focus, and we introduce four methods, namely \textbf{NAIM3L}, \textbf{iMVWL}, \textbf{DICNet}, \textbf{LMVCAT} and \textbf{MTD}, which aim to address the IMVMLC challenge.

\textbf{Evaluation}: Similar to previous works \cite{Alpher33,Alpher35,Alpher39}, we select six metrics to measure these methods, i.e., Average Precision (\textbf{AP}), Hamming Loss (\textbf{HL}), Ranking Loss (\textbf{RL}), adapted area under curve (\textbf{AUC}), OneError (\textbf{OE}), and Coverage (\textbf{Cov}). Specially, to intuitively observe the difference in performance, we report results concerning \textbf{1-HL} and \textbf{1-RL}. It should be noted that the higher the values of the four metrics, the better the performance is.

\begin{table*}[ht]
  \caption{The performance of different methods on various datasets with $50\%$ missing-view rate, $50\%$ missing-label rate and $70\%$ training samples.}\resizebox{0.95\textwidth}{!}{
  \label{tab:commands}
  \begin{tabular}{ccccccccc}
    \toprule
    DATA & METRIC & iMvWL & NAIM3L & CDMM &  DICNet & LMVCAT & MTD & OURS \\
    \midrule
    & AP & 0.283$\pm$0.011 & 0.309$\pm$0.004 & 0.309$\pm$0.004 & 0.381$\pm$0.004 & 0.382$\pm$0.004 & 0.415$\pm$0.008 &\textbf{0.425$\pm$0.006} \\
    & 1-HL & 0.978$\pm$0.000 & 0.987$\pm$0.000 & 0.987$\pm$0.000 & 0.988$\pm$0.000 & 0.988$\pm$0.000 & 0.988$\pm$0.000 & \textbf{0.988$\pm$0.002} \\ 
    & 1-RL & 0.865$\pm$0.005 & 0.878$\pm$0.002 & 0.884$\pm$0.003 & 0.882$\pm$0.004 & 0.880$\pm$0.002 & 0.893$\pm$0.004 & \textbf{0.898$\pm$0.003} \\
    Corel5k & AUC &  0.868$\pm$0.005 & 0.881$\pm$0.002 & 0.888$\pm$0.003 & 0.884$\pm$0.004 & 0.883$\pm$0.002  & 0.896$\pm$0.004 & \textbf{0.903$\pm$0.004} \\
    & OE &  0.689$\pm$0.015 & 0.650$\pm$0.009 & 0.590$\pm$0.007 & 0.532$\pm$0.007 & 0.547$\pm$0.006 & 0.509$\pm$0.012 & \textbf{0.494$\pm$0.010} \\
    & Cov & 0.298$\pm$0.008 & 0.275$\pm$0.005 & 0.277$\pm$0.007 & 0.273$\pm$0.011 & 0.273$\pm$0.006 & 0.251$\pm$0.009 & \textbf{0.246$\pm$0.007} \\
    \midrule
    & AP & 0.437$\pm$0.018 & 0.488$\pm$0.003 & 0.508$\pm$0.005 & 0.505$\pm$0.012 & 0.519$\pm$0.005 & 0.551$\pm$0.004 & \textbf{0.558$\pm$0.004} \\
    & 1-HL & 0.882$\pm$0.004 & 0.928$\pm$0.001 & 0.931$\pm$0.001 & 0.929$\pm$0.001 & 0.924$\pm$0.003 & 0.932$\pm$0.001 & \textbf{0.934$\pm$0.000} \\
    & 1-RL & 0.736$\pm$0.015 & 0.783$\pm$0.001 & 0.812$\pm$0.004 & 0.783$\pm$0.008 & 0.811$\pm$0.004 & 0.831$\pm$0.003 & \textbf{0.837$\pm$0.003} \\
    Pascal07 & AUC & 0.767$\pm$0.015 & 0.811$\pm$0.001 & 0.838$\pm$0.003 & 0.809$\pm$0.006 & 0.834$\pm$0.004 & 0.851$\pm$0.003 & \textbf{0.857$\pm$0.004} \\
    & OE & 0.638$\pm$0.023 & 0.579$\pm$0.006 & 0.581$\pm$0.008 & 0.573$\pm$0.015 & 0.579$\pm$0.006 & 0.541$\pm$0.004 & \textbf{0.548$\pm$0.006} \\
    & Cov & 0.323$\pm$0.015 & 0.273$\pm$0.002 & 0.241$\pm$0.003 & 0.269$\pm$0.006 & 0.237$\pm$0.005 & 0.216$\pm$0.004 & \textbf{0.212$\pm$0.007} \\
    \midrule
    & AP & 0.244$\pm$0.005 & 0.246$\pm$0.002 & 0.289$\pm$0.003 & 0.297$\pm$0.002 & 0.294$\pm$0.004 & 0.306$\pm$0.003 & \textbf{0.317$\pm$0.003} \\
    & 1-HL & 0.972$\pm$0.000 & 0.983$\pm$0.000 & 0.983$\pm$0.000 & 0.983$\pm$0.000 & 0.982$\pm$0.000 & 0.983$\pm$0.000 & \textbf{0.983$\pm$0.000} \\
    & 1-RL & 0.808$\pm$0.002 & 0.818$\pm$0.002 & 0.832$\pm$0.001 & 0.832$\pm$0.001 & 0.828$\pm$0.002 & 0.837$\pm$0.002 & \textbf{0.849$\pm$0.002} \\
    ESPGame & AUC & 0.813$\pm$0.002 & 0.824$\pm$0.002 & 0.836$\pm$0.001 & 0.836$\pm$0.001 & 0.833$\pm$0.002 & 0.842$\pm$0.002 & \textbf{0.855$\pm$0.003} \\
    & OE & 0.657$\pm$0.013 & 0.661$\pm$0.003 & 0.604$\pm$0.005 & 0.561$\pm$0.007 & 0.566$\pm$0.009 & 0.553$\pm$0.009 & \textbf{0.542$\pm$0.007} \\
    & Cov & 0.452$\pm$0.004 & 0.429$\pm$0.003 & 0.426$\pm$0.004 & 0.407$\pm$0.003 & 0.410$\pm$0.004 & 0.398$\pm$0.004 & \textbf{0.370$\pm$0.004} \\
    \midrule
    & AP & 0.237$\pm$0.003 & 0.261$\pm$0.001 & 0.305$\pm$0.004 & 0.323$\pm$0.001 & 0.317$\pm$0.003 & 0.332$\pm$0.003 & \textbf{0.339$\pm$0.003} \\
    & 1-HL & 0.969$\pm$0.000 & 0.980$\pm$0.000 & 0.981$\pm$0.000 & 0.981$\pm$0.000 & 0.980$\pm$0.000 & 0.981$\pm$0.000 & \textbf{0.981$\pm$0.002} \\
    & 1-RL & 0.833$\pm$0.002 & 0.848$\pm$0.001 & 0.862$\pm$0.002 & 0.873$\pm$0.001 & 0.870$\pm$0.001 & 0.875$\pm$0.002 & \textbf{0.888$\pm$0.004} \\
    IAPRTC12 & AUC &  0.835$\pm$0.001 & 0.850$\pm$0.001 & 0.864$\pm$0.002 & 0.874$\pm$0.000 & 0.876$\pm$0.001 & 0.876$\pm$0.004 & \textbf{0.889$\pm$0.004} \\
    & OE & 0.648$\pm$0.008 & 0.610$\pm$0.005 & 0.568$\pm$0.008 & 0.532$\pm$0.002 & 0.557$\pm$0.005 & 0.533$\pm$0.004 & \textbf{0.525$\pm$0.003} \\
    & Cov & 0.436$\pm$0.005 & 0.408$\pm$0.004 & 0.403$\pm$0.004 & 0.351$\pm$0.001 & 0.352$\pm$0.003 & 0.351$\pm$0.004 & \textbf{0.322$\pm$0.007} \\
    \midrule
    & AP &  0.490$\pm$0.012 & 0.551$\pm$0.002 & 0.570$\pm$0.002 & 0.589$\pm$0.005 & 0.594$\pm$0.005 & 0.607$\pm$0.004 & \textbf{0.619$\pm$0.004} \\
    & 1-HL & 0.839$\pm$0.002 & 0.882$\pm$0.001 & 0.886$\pm$0.001 & 0.888$\pm$0.002 & 0.882$\pm$0.002 & 0.891$\pm$0.001 & \textbf{0.894$\pm$0.001} \\
    & 1-RL &  0.803$\pm$0.008 & 0.844$\pm$0.001 & 0.856$\pm$0.001 &0.865$\pm$0.003 & 0.863$\pm$0.004 & 0.875$\pm$0.002 & \textbf{0.881$\pm$0.001} \\
    Mirflickr & AUC &  0.787$\pm$0.012 & 0.837$\pm$0.001 & 0.846$\pm$0.001 &  0.853$\pm$0.003 & 0.849$\pm$0.004 & 0.862$\pm$0.002 & \textbf{0.864$\pm$0.001} \\
    & OE &  0.489$\pm$0.022 & 0.415$\pm$0.003 & 0.369$\pm$0.004 & 
    0.358$\pm$0.008 & 0.363$\pm$0.007 & 0.345$\pm$0.004 & \textbf{0.332$\pm$0.004}\\
    & Cov & 0.428$\pm$0.013 & 0.369$\pm$0.002 & 0.360$\pm$0.001 & 0.333$\pm$0.003 & 0.348$\pm$0.007 & 0.324$\pm$0.004 & \textbf{0.313$\pm$0.003} \\
    \bottomrule
  \end{tabular}
  }
\end{table*}

\begin{figure}
\includegraphics[width=\textwidth]{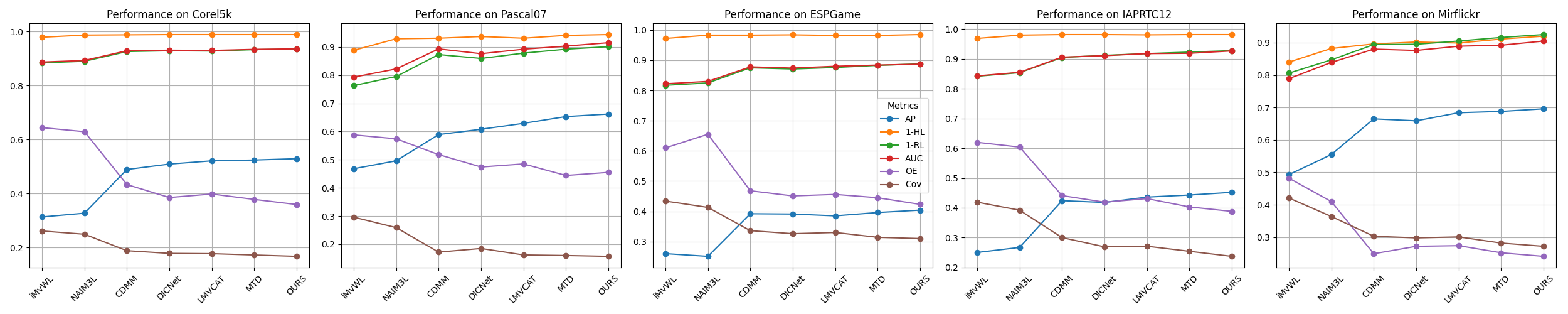}
\caption{The performance of different methods on various datasets with full
views, full labels and $70\%$ training samples.}
\end{figure}

\begin{figure}[!t]
\centering
\subfigure[different missing-view ratios]{\includegraphics[width=1.8in,height=1.5in]{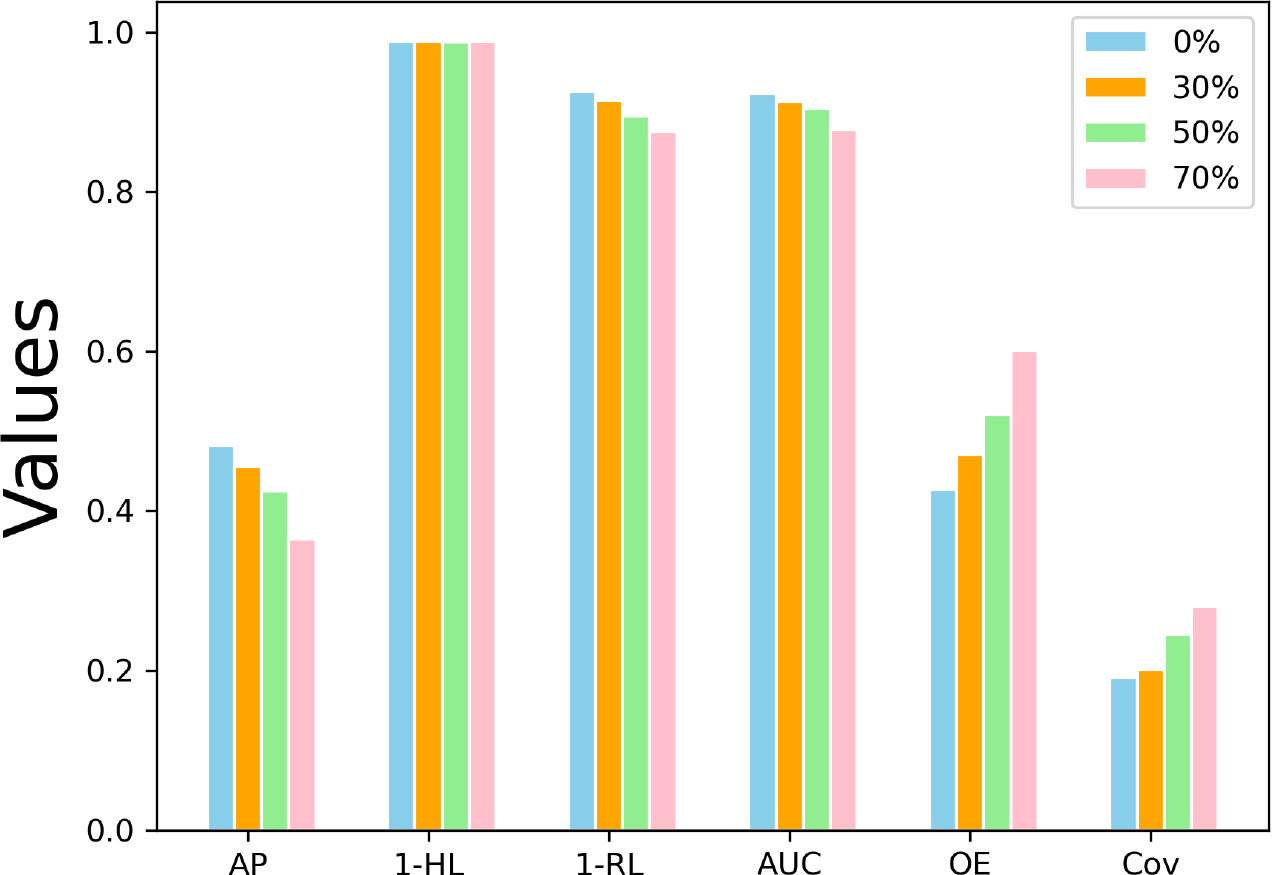}}
\hfil
\subfigure[different missing-label ratios]{\includegraphics[width=1.8in,height=1.5in]{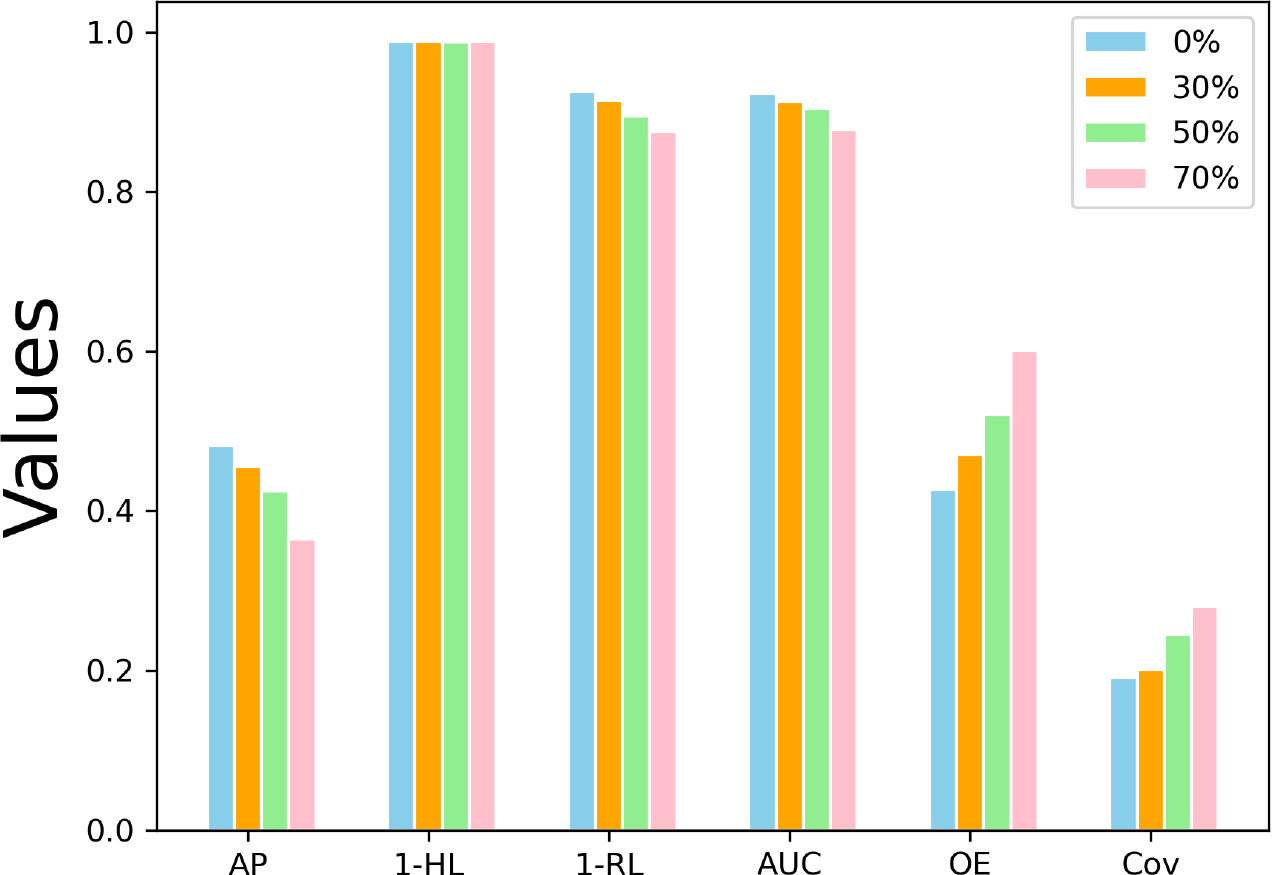}}
\caption{Experimental results on the Corel5k dataset: (a) different missing-view ratios and a 50\% missing-label ratio and (b) different missing-label ratios and a 50\% missing-view ratio.}
\end{figure}

\subsection{Experimental Results and Analysis}
In this section, we compare the performance of our proposed framework by comparing it with six competitive methods on the aforementioned five datasets with scenarios of $50\%$ missing-view rate, $50\%$ missing-label rate and $70\%$ training samples, as shown in Table 1. To further explore the impacts of missing views and missing labels, we conducted experiments on the Corel5k dataset and Fig 2 shows the detail of our results related to different missing views and missing label ratios. According to the statistical results shown in Table 2 and Fig.2, it is evident to make the following observations:
\begin{itemize}
\item As shown in Table 1 and Fig 2, our method outperforms all models on all metrics of both five datasets, which fully demonstrates the effectiveness of our proposed method. Moreover, as shown in Fig.2, the increase in the missing ratio of views and labels leads to worse performance. Moreover, the impact of missing views is more significant than missing labels.
\item Compared with traditional methods, the DNN-based methods show significant advantages to iMvMLC problems. This indicates the necessity to design dedicated methods for missing problems.
\end{itemize}
Furthermore, in order to evaluate the effectiveness of our decoupled dual contrastive framework, we calculate the mean feature of all available samples in each channel. Fig.4(a)-Fig.4(d) show the average feature channel similarity heatmaps of all channels at epochs 0, 20, 40, and 60 on the Corel5k dataset, where half of the views and labels are missing. As can be seen from the figure, at the beginning of training, the similarity between different channels is relatively small. However, as training progresses, the similarity between instance pairs on the first v shared channels increases rapidly. This observation suggests that our dual contrast framework can fully extract high-quality shared information from multi-view data.

\begin{figure}[!t]
\small
\centering
\subfigure[Corel5k]{\includegraphics[width=1.5in,height=1.2in]{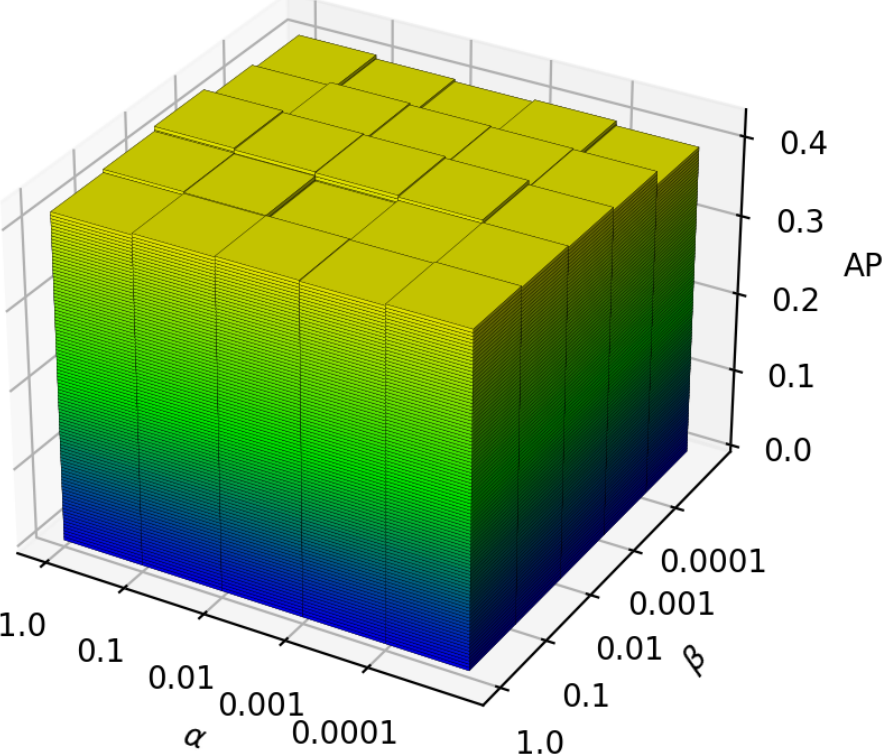}}
\subfigure[Pascal07]{\includegraphics[width=1.5in,height=1.2in]{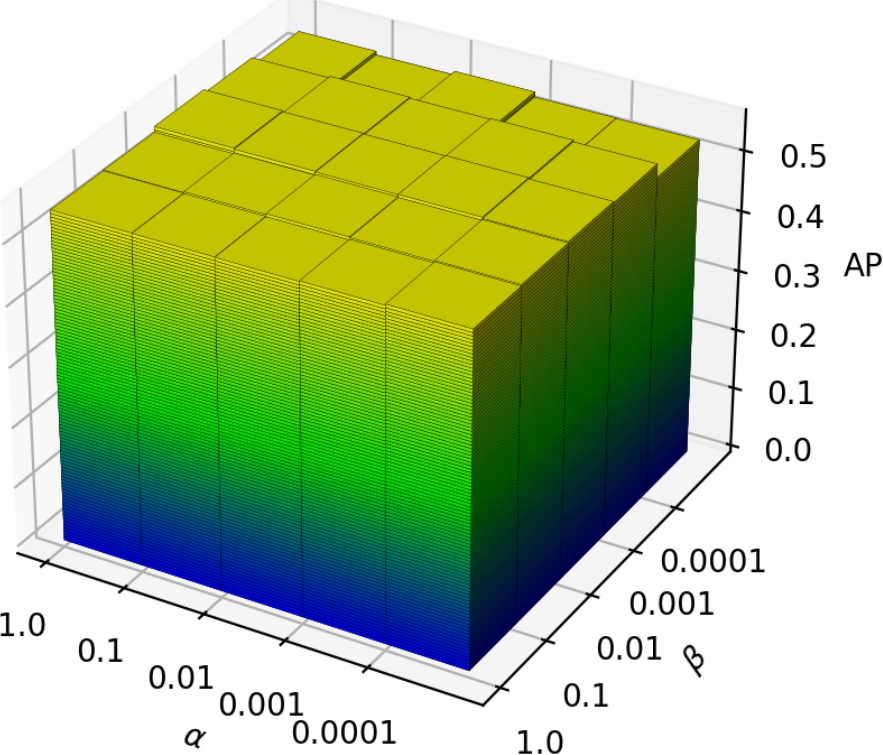}}
\hfil
\subfigure[Corel5k and Pascal07]{\includegraphics[width=1.5in,height=1.2in]{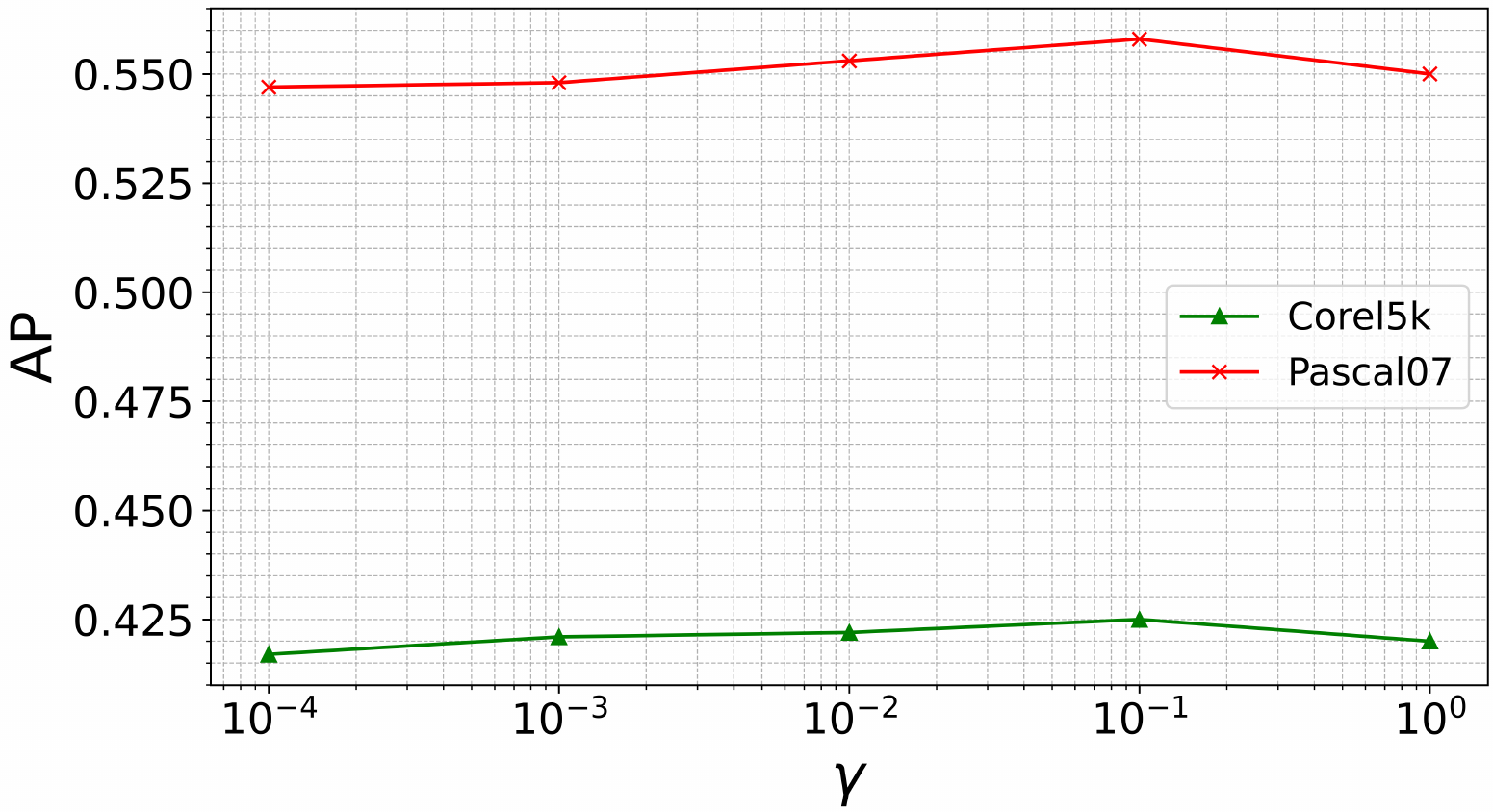}}
\caption{The AP values for hyper-parameters $\alpha$ and $\beta$ on the Corel5k (Fig. 3a) and Pascal07 (Fig. 3b) datasets; AP values for $\gamma$ on both Corel5k and Pascal07 datasets (Fig. 3c). Both datasets contain 50\% available views and labels, with a 70\% training sample rate.}
\end{figure}

\begin{figure}[!t]
\small
\centering
\subfigure[Epoch0]{\includegraphics[width=2.3in,height=1.6in]{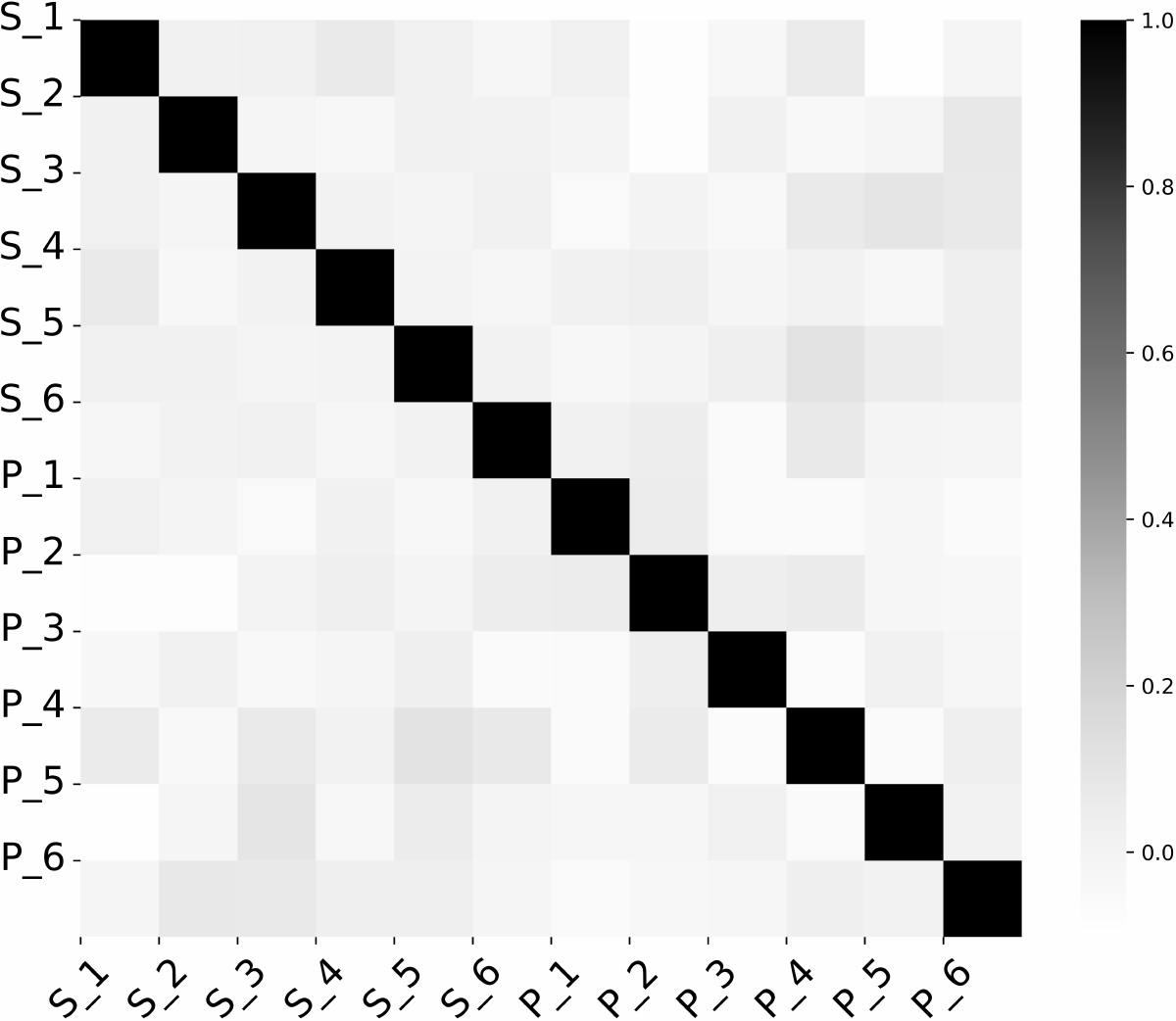}}
\subfigure[Epoch20]{\includegraphics[width=2.3in,height=1.6in]{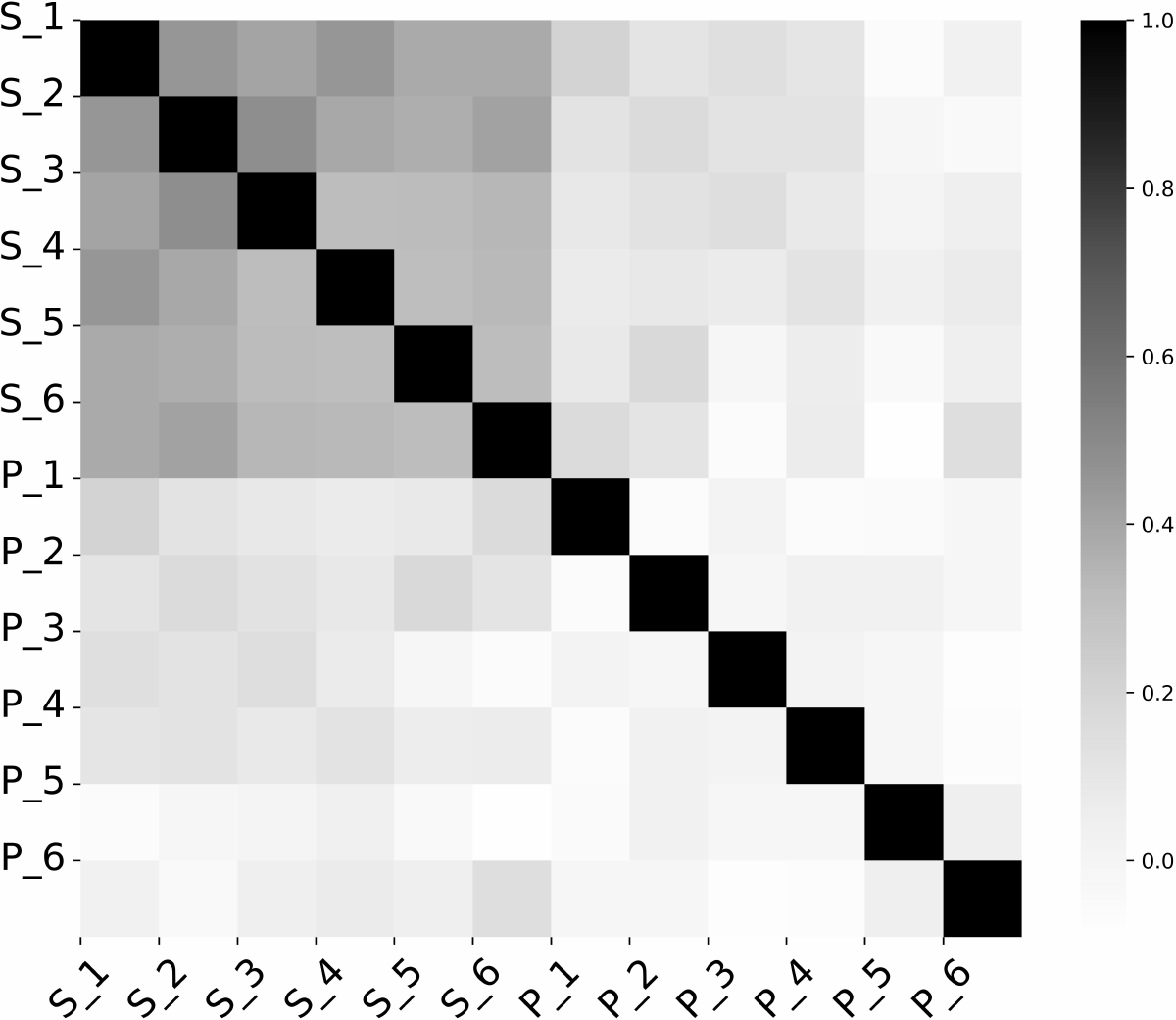}}
\subfigure[Epoch40]{\includegraphics[width=2.3in,height=1.6in]{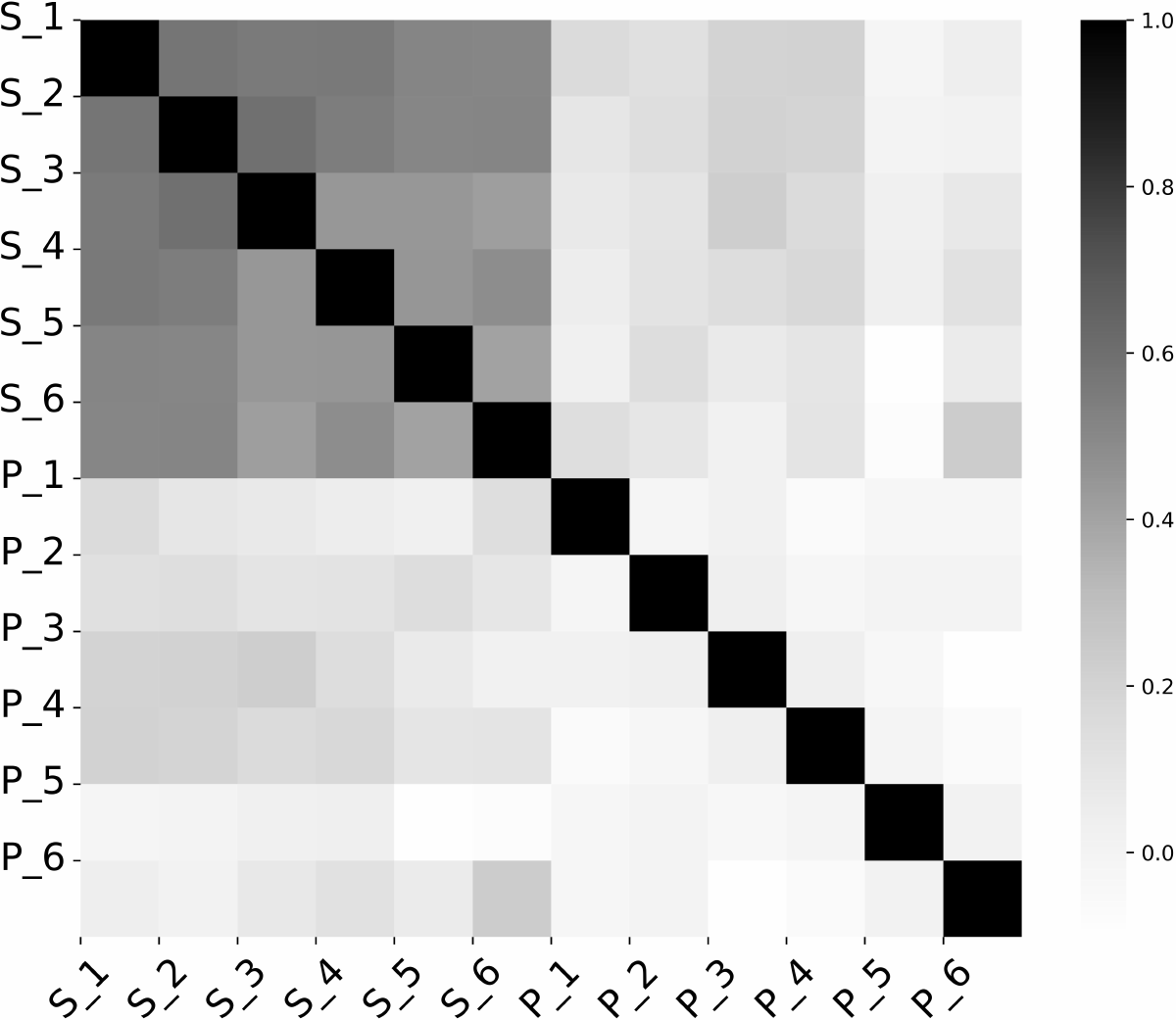}}
\subfigure[Epoch60]{\includegraphics[width=2.3in,height=1.6in]{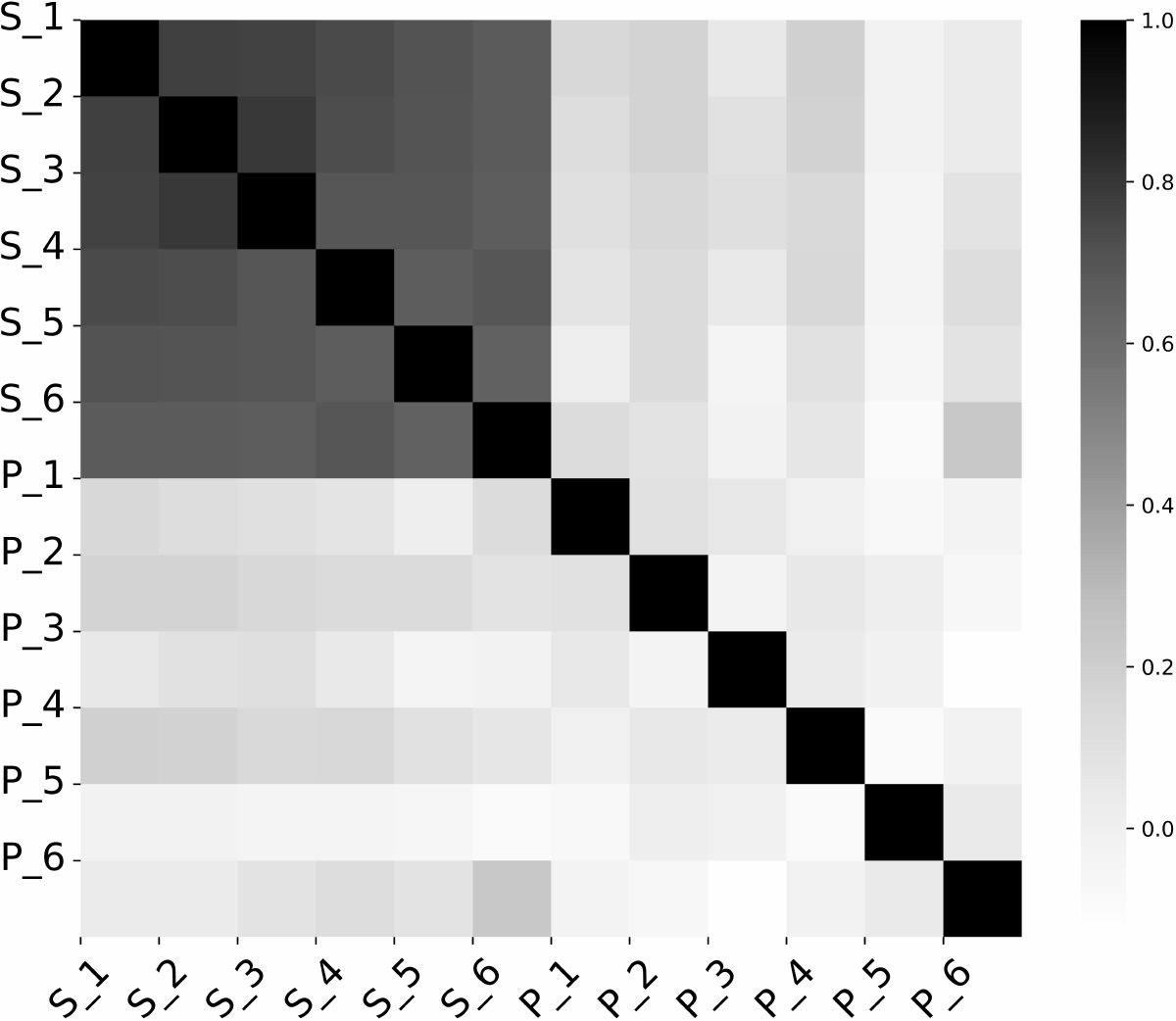}}
\caption{The following figures present a channel similarity heatmap of the average feature of all samples in different channels on the Corel5k dataset, where half of the views and labels are absent. The six groups of features, labeled $S_1-S_6$ and $P_1-P_6$, represent the shared features and view-private features on the six views, respectively. As the number of training epochs increases, the similarity of shared features between views gradually increases, while the similarity between shared features and private features gradually decreases.}
\end{figure}

\begin{table}[!t]
\footnotesize
\centering
\caption{The ablation experiment results on the Corel5k and Mirflickr datasets with $50\%$ missing-view rate, $50\%$ missing-label rate and $70\%$ training samples.}
\label{my-label}
\resizebox{0.50\linewidth}{!}{
\begin{tabular}{@{}cccccc|cc@{}}
\toprule
Backbone & $L_{S}$ & $L_{l}$ & $L_{r}$ & \multicolumn{2}{c|}{Core15k} & \multicolumn{2}{c}{Mirflickr} \\ 
\cmidrule(lr){5-6} \cmidrule(lr){7-8}
         &          &           &           & AP    & AUC   & AP    & AUC   \\
\midrule
\checkmark &          &           &           & 0.389 & 0.888 & 0.603 & 0.861 \\
\checkmark & \checkmark &         &           & 0.402 & 0.895 & 0.611 & 0.864 \\
\checkmark &          & \checkmark &         & 0.406 & 0.897 & 0.612 & 0.864 \\
\checkmark &          &           & \checkmark & 0.398 & 0.893 & 0.602 & 0.861 \\
\checkmark &   \checkmark       &     \checkmark      &           &   0.407    &  0.898     &    0.615   &   0.867    \\
\checkmark & \checkmark &  &   \checkmark    & 0.407 & 0.899 & 0.612 & 0.863 \\
\checkmark &          & \checkmark & \checkmark & 0.406 & 0.900 & 0.610 & 0.862 \\
\checkmark & \checkmark & \checkmark & \checkmark &\textbf{0.425} & \textbf{0.903} & \textbf{0.619} & \textbf{0.864} \\
\bottomrule
\end{tabular}
}
\end{table}

\subsection{Hyperparameters Analysis}
In our DCL model, there are three hyper-parameters, i.e., $\alpha, \beta$ and $\gamma $ that need to be set before training. The sensitivity of the model was explored by varying the values and reporting the corresponding AP values on the Corel5k and Pascal07 datasets, respectively. From Figure.3a and Figure. 3b, we can see that in the condition of $50\%$ missing views, $50\%$ missing labels and $70\%$ training samples, the optimal ranges of $\alpha$ and $\beta$ on Corel5k and Pascal07 are all [0.0001,1]. From Figure. 3c, we can observe that our model is not sensitive to $\gamma$ and we set $\gamma$ as $1e-1$ for all datasets.

\subsection{Ablation Study}
To find out the effectiveness of various parts of our model, we perform ablation experiments on Corel5k and Mirflickr datasets with $50\%$ missing views, $50\%$ missing labels and $70\%$ training samples. We removed $L_{s} L_{l} L_{r}$, respectively. it is possible to obtain some key conclusions from Table 2: (i) Each component of our DCL plays an important role and contributes to the overall improvement of multi-label classification. (ii) The most effective improvement is our dual-level contrastive loss.

\section{Conclusion}
In this paper, we have proposed a dual-level contrastive learning framework (DCL) for the IMVMLC task. For each view, our DCL decouples its features into two channels for consistency and complementarity learning, separately. The application of indicator matrices effectively avoids the missing-view and weak label challenge. With low-level features extracted by S-P encoders, on the one hand, a shared MLP and classifier are introduced to map shared information into high-level space, which reduces the conflict between the reconstruction objective and consensus objective. On the other hand, the private information is constrained to make sure that the learned embedding features preserve structure information among samples. Last but not least, considering missing data may degrade the performance, the masked input strategy is designed to reduce heavy spatial redundancy. Extensive experiments on five popular datasets demonstrate the superiority of our method. 

\clearpage

%
%
%
%

\bibliography{egbib}

\begin{thebibliography}{10}
\providecommand{\url}[1]{\texttt{#1}}
\providecommand{\urlprefix}{URL }
\providecommand{\doi}[1]{https://doi.org/#1}

\bibitem{Alpher11}
Chaudhuri, K., Kakade, S.M., Livescu, K., Sridharan, K.: Multi-view clustering via canonical correlation analysis. In: Proceedings of the 26th annual international conference on machine learning. pp. 129--136 (2009)

\bibitem{Alpher38}
Che, X., Chen, D., Mi, J.: A novel approach for learning label correlation with application to feature selection of multi-label data. Information Sciences  \textbf{512},  795--812 (2020)

\bibitem{Alpher12}
Chen, N., Zhu, J., Xing, E.: Predictive subspace learning for multi-view data: a large margin approach. Advances in neural information processing systems  \textbf{23} (2010)

\bibitem{Diao_2025_WACV}
Diao, X., Cheng, M., Barrios, W., Jin, S.: Ft2tf: First-person statement text-to-talking face generation. In: Proceedings of the Winter Conference on Applications of Computer Vision (WACV). pp. 4821--4830 (February 2025)

\bibitem{diao2025learning}
Diao, X., Yang, T., Zhang, C., Wu, W., Cheng, M., Gui, J.: Learning sparsity for effective and efficient music performance question answering. arXiv preprint arXiv:2506.01319  (2025)

\bibitem{diao2025soundmind}
Diao, X., Zhang, C., Kong, K., Wu, W., Ma, C., Ouyang, Z., Qing, P., Vosoughi, S., Gui, J.: Soundmind: Rl-incentivized logic reasoning for audio-language models. arXiv preprint arXiv:2506.12935  (2025)

\bibitem{diao-etal-2024-learning}
Diao, X., Zhang, C., Wu, T., Cheng, M., Ouyang, Z., Wu, W., Gui, J.: Learning musical representations for music performance question answering. In: Findings of the Association for Computational Linguistics: EMNLP 2024 (2024)

\bibitem{diao2025temporal}
Diao, X., Zhang, C., Wu, W., Ouyang, Z., Qing, P., Cheng, M., Vosoughi, S., Gui, J.: Temporal working memory: Query-guided segment refinement for enhanced multimodal understanding. arXiv preprint arXiv:2502.06020  (2025)

\bibitem{Alpher16}
Ding, G., Guo, Y., Zhou, J.: Collective matrix factorization hashing for multimodal data. In: Proceedings of the IEEE conference on computer vision and pattern recognition. pp. 2075--2082 (2014)

\bibitem{Alpher18}
Frome, A., Corrado, G.S., Shlens, J., Bengio, S., Dean, J., Ranzato, M., Mikolov, T.: Devise: A deep visual-semantic embedding model. Advances in neural information processing systems  \textbf{26} (2013)

\bibitem{Alpher15}
Gao, S., Yu, Z., Jin, T., Yin, M.: Multi-view low-rank matrix factorization using multiple manifold regularization. Neurocomputing  \textbf{335},  143--152 (2019)

\bibitem{Alpher07}
Guillaumin, M., Verbeek, J., Schmid, C.: Multimodal semi-supervised learning for image classification. In: 2010 IEEE Computer society conference on computer vision and pattern recognition. pp. 902--909. IEEE (2010)

\bibitem{Alpher08}
Han, Z., Zhang, C., Fu, H., Zhou, J.T.: Trusted multi-view classification. In: International Conference on Learning Representations (2020)

\bibitem{Alpher37}
Huang, S.J., Zhou, Z.H.: Multi-label learning by exploiting label correlations locally. In: Proceedings of the AAAI Conference on Artificial Intelligence. vol.~26, pp. 949--955 (2012)

\bibitem{Alpher36}
Khosla, P., Teterwak, P., Wang, C., Sarna, A., Tian, Y., Isola, P., Maschinot, A., Liu, C., Krishnan, D.: Supervised contrastive learning. Advances in neural information processing systems  \textbf{33},  18661--18673 (2020)

\bibitem{Alpher35}
Li, X., Chen, S.: A concise yet effective model for non-aligned incomplete multi-view and missing multi-label learning. IEEE Transactions on Pattern Analysis and Machine Intelligence  \textbf{44}(10),  5918--5932 (2021)

\bibitem{Alpher10}
Lin, Y., Gou, Y., Liu, Z., Li, B., Lv, J., Peng, X.: Completer: Incomplete multi-view clustering via contrastive prediction. In: Proceedings of the IEEE/CVF conference on computer vision and pattern recognition. pp. 11174--11183 (2021)

\bibitem{Alpher33}
Liu, C., Wen, J., Liu, Y., Huang, C., Wu, Z., Luo, X., Xu, Y.: Masked two-channel decoupling framework for incomplete multi-view weak multi-label learning. Advances in Neural Information Processing Systems  \textbf{36} (2024)

\bibitem{Alpher31}
Liu, C., Wen, J., Luo, X., Huang, C., Wu, Z., Xu, Y.: Dicnet: Deep instance-level contrastive network for double incomplete multi-view multi-label classification. In: Proceedings of the AAAI conference on artificial intelligence. vol.~37, pp. 8807--8815 (2023)

\bibitem{Alpher42}
Liu, C., Wen, J., Luo, X., Xu, Y.: Incomplete multi-view multi-label learning via label-guided masked view- and category-aware transformers. In: Proceedings of the AAAI Conference on Artificial Intelligence. vol.~37, pp. 8816--8824 (2023)

\bibitem{Alpher28}
Liu, C., Wen, J., Luo, X., Xu, Y.: Incomplete multi-view multi-label learning via label-guided masked view-and category-aware transformers. In: Proceedings of the AAAI Conference on Artificial Intelligence. vol.~37, pp. 8816--8824 (2023)

\bibitem{Alpher24}
Liu, M., Luo, Y., Tao, D., Xu, C., Wen, Y.: Low-rank multi-view learning in matrix completion for multi-label image classification. In: Proceedings of the AAAI conference on artificial intelligence. vol.~29 (2015)

\bibitem{Alpher32}
Liu, X., Sun, L., Feng, S.: Incomplete multi-view partial multi-label learning. Applied Intelligence  \textbf{52}(3),  3289--3302 (2022)

\bibitem{Alpher05}
Poria, S., Majumder, N., Hazarika, D., Cambria, E., Gelbukh, A., Hussain, A.: Multimodal sentiment analysis: Addressing key issues and setting up the baselines. IEEE Intelligent Systems  \textbf{33}(6),  17--25 (2018)

\bibitem{Alpher09}
Seeland, M., M{\"a}der, P.: Multi-view classification with convolutional neural networks. Plos one  \textbf{16}(1),  e0245230 (2021)

\bibitem{Alpher39}
Tan, Q., Yu, G., Domeniconi, C., Wang, J., Zhang, Z.: Incomplete multi-view weak-label learning. In: Ijcai. pp. 2703--2709 (2018)

\bibitem{Alpher04}
Tran~Gia, B., Bui Cong~Khanh, T., Tran~Nhat, K., Luu~Trung, K., Tran~Doan, T., Le~Tran~Trong, K., Do, T., Duc~Ngo, T.: Integrating multiple models for effective video retrieval and multi-stage search. In: Proceedings of the 12th International Symposium on Information and Communication Technology. pp. 1003--1010 (2023)

\bibitem{Alpher06}
Tsai, D., Jing, Y., Liu, Y., Rowley, H.A., Ioffe, S., Rehg, J.M.: Large-scale image annotation using visual synset. In: 2011 International Conference on Computer Vision. pp. 611--618. IEEE (2011)

\bibitem{Alpher20}
Wang, X., Bian, W., Tao, D.: Grassmannian regularized structured multi-view embedding for image classification. IEEE Transactions on Image Processing  \textbf{22}(7),  2646--2660 (2013). \doi{10.1109/TIP.2013.2255300}

\bibitem{Alpher30}
Wen, J., Liu, C., Deng, S., Liu, Y., Fei, L., Yan, K., Xu, Y.: Deep double incomplete multi-view multi-label learning with incomplete labels and missing views. IEEE Transactions on Neural Networks and Learning Systems  (2023)

\bibitem{Alpher21}
Wu, X., Chen, Q.G., Hu, Y., Wang, D., Chang, X., Wang, X., Zhang, M.L.: Multi-view multi-label learning with view-specific information extraction. In: IJCAI. pp. 3884--3890 (2019)

\bibitem{Alpher34}
Xu, J., Tang, H., Ren, Y., Peng, L., Zhu, X., He, L.: Multi-level feature learning for contrastive multi-view clustering. In: Proceedings of the IEEE/CVF conference on computer vision and pattern recognition. pp. 16051--16060 (2022)

\bibitem{Alpher19}
Xu, J., Li, W., Liu, X., Zhang, D., Liu, J., Han, J.: Deep embedded complementary and interactive information for multi-view classification. In: Proceedings of the AAAI conference on artificial intelligence. vol.~34, pp. 6494--6501 (2020)

\bibitem{Alpher26}
Zhao, D., Gao, Q., Lu, Y., Sun, D., Cheng, Y.: Consistency and diversity neural network multi-view multi-label learning. Knowledge-Based Systems  \textbf{218},  106841 (2021)

\bibitem{Alpher01}
Zhao, J., Xie, X., Xu, X., Sun, S.: Multi-view learning overview: Recent progress and new challenges. Inf. Fusion  \textbf{38},  43--54 (2017), \url{https://api.semanticscholar.org/CorpusID:35944072}

\bibitem{Alpher22}
Zhao, X., Chen, Y., Liu, S., Tang, B.: Shared-private memory networks for multimodal sentiment analysis. IEEE Transactions on Affective Computing  (2022)

\bibitem{Alpher41}
Zhu, C., Liu, Y., Miao, D., Dong, Y., Pedrycz, W.: Within- cross- consensus-view representation-based multi-view multi-label learning with incomplete data. Neurocomputing  \textbf{557},  126729 (2023), \url{https://api.semanticscholar.org/CorpusID:261193108}

\end{thebibliography}
\bibliographystyle{splncs04}
\end{document}